\documentclass{article}

\usepackage{xcolor}
\usepackage{arxiv}
\usepackage[utf8]{inputenc} 
\usepackage[T1]{fontenc}    
\usepackage{hyperref}       
\usepackage{url}            
\usepackage{booktabs}       
\usepackage{amsfonts}       
\usepackage{nicefrac}       
\usepackage{microtype}      
\usepackage{lipsum}		
\usepackage{graphicx}
\usepackage{natbib}
\setcitestyle{numbers}
\usepackage{doi}
\setcitestyle{square}

\usepackage{caption} 
\usepackage{graphicx}
\usepackage{amsmath,amssymb,amsthm}
\usepackage{geometry} 

\newtheorem{remark}{Remark}
\newtheorem{theorem}{Theorem}

\title{Generating synthetic data for neural operators}

\author{ Erisa Hasani
\thanks{Corresponding author} \\
	Department of Mathematics\\
	University of Texas at Austin\\
	\texttt{ehasani@utexas.edu} \\
	\And
	\hspace{1mm}Rachel A. Ward \\
	Department of Mathematics\\
	University of Texas at Austin\\
         Microsoft Research \\
	\texttt{rward@math.utexas.edu} \\
}

\date{}

\renewcommand{\headeright}{}
\renewcommand{\undertitle}{}

\begin{document}
\maketitle
\begin{abstract}
 Recent advances in the literature show promising potential of deep learning methods, particularly neural operators, in obtaining numerical solutions to partial differential equations (PDEs) beyond the reach of current numerical solvers. However, existing data-driven approaches often rely on training data produced by numerical PDE solvers (e.g., finite difference or finite element methods). We introduce a "backward" data generation method that avoids solving the PDE numerically: by randomly sampling candidate solutions $u_j$ from the appropriate solution space (e.g., $H_0^1(\Omega)$), we compute the corresponding right-hand side $f_j$ directly from the equation by differentiation. This produces training pairs ${(f_j, u_j)}$ by computing derivatives rather than solving a PDE numerically for each data point, enabling fast, large-scale data generation consisting of exact solutions. Experiments indicate that models trained on this synthetic data generalize well when tested on data produced by standard solvers. While the idea is simple, we hope this method will expand the potential of neural PDE solvers that do not rely on classical numerical solvers to generate their data. 
\end{abstract}


\section{Introduction}
The use of deep learning to obtain numerical solutions to PDE problems beyond the reach of classical solvers shows promise in revolutionizing science and technology. Deep learning-based methods have overcome many challenges that classical numerical methods suffer from, among which are the curse of dimensionality and grid dependence. 

Methods that attempt to solve PDE problems using deep learning can be split into two main classes: those that solve an instance of a PDE problem by directly approximating the solution (e.g. \cite{DeepRitz}, \cite{DGM}, \cite{PINNs}, \cite{DBLP:journals/corr/abs-2101-04879}, \cite{Zhang_2022}, \cite{DBLP:journals/ijon/BilginVM22}), and those that consider solutions to a family of PDE problems, also known in the literature as parametric PDEs, through operator learning.  In the operator learning approach, the goal is to approximate the solution operator that maps input functions to the unknown solution (e.g. \cite{DeepONet}, \cite{modelReduction}, \cite{graph_no_2020}, \cite{NeuralOperator}, \cite{CNO}, \cite{penwarden2023metalearning}). In this paper, we focus on the second class, where we seek solutions to a class of PDE problems instead of an instance. Although the approach we describe is general, we focus on the Fourier Neural Operator (FNO) \cite{NeuralOperator}, which is a state-of-the-art neural operator learning method at the time that this paper is being written.  We stress that our method is independent of the particular neural operator learning architecture and should remain applicable as a synthetic data generation plug-in as the state-of-the-art architecture evolves.

To the best of our knowledge, classical numerical methods, such as finite differences, finite element \cite{THOMEE20011}, pseudo-spectral methods, or other variants, have been used to obtain data for training purposes in operator learning. In particular, some works have used finite difference schemes (e.g. \cite{DeepONet}, \cite{modelReduction}, \cite{DBLP:conf/nips/LiKALSBA20}, \cite{DBLP:journals/corr/abs-2202-08942}, \cite{DBLP:journals/corr/abs-2204-11127}, \cite{DBLP:journals/corr/abs-2206-09418}, \cite{CNO}, \cite{DBLP:conf/icml/MolinaroYEM23}). In other works, data has been generated by constructing examples that have a closed-form explicit solution or by using schemes such as finite element, pseudo-spectral schemes, fourth-order Runge-Kutta, forward Euler, etc. (e.g. \cite{DBLP:conf/icml/LongLMD18}, \cite{DBLP:journals/corr/abs-2103-10974}, \cite{DBLP:journals/corr/abs-2205-10573}, \cite{DBLP:conf/nips/SeidmanKPP22}, \cite{DBLP:journals/corr/abs-2207-05209}, \cite{DBLP:journals/corr/abs-2212-04689}, \cite{DBLP:journals/corr/abs-2201-11967}, \cite{DBLP:conf/iclr/TranMXO23}, \cite{DBLP:journals/corr/abs-2303-10528}, \cite{DBLP:conf/icml/VadeboncoeurKPC23}, \cite{DBLP:journals/corr/abs-2305-08573}, \cite{DBLP:journals/corr/abs-2307-15034}). While these works are a strong proof-of-concept for neural operators, it is critical to not rely completely on using classical numerical solvers to generate training data for neural operator learning if we want to develop neural operators as general-purpose PDE solvers \emph{beyond} the reach of classical numerical solvers.

 {\bf Our approach.} Our approach is conceptually simple: suppose we want to train a neural network to learn solutions to a parameterized class of PDE problems of the form \eqref{1.1}. If we know that the solution for any value of the parameter belongs to a Sobolev space which has an explicit orthonormal basis of eigenfunctions and associated eigenvalues, we can generate a large number of synthetic training functions $\{u_{a_j}^{k}\}_{j,k}$ in the space as random linear combinations of the first so many eigenfunctions, scaled by the corresponding eigenvalues (see Section \ref{data_gen} for more details). We can efficiently generate corresponding right-hand side functions $f_{a_j}^k$ by computing derivatives, $-L_{a_j} u_{a_j}^k = f_{a_j}^k$.  We then use training data $(f_{a_j}^k, a_j, u_{a_j}^k)_{j=1}^N$ to train a neural operator to learn the class of PDEs. 

 The general concept of first generating the “unknown” function and then substituting it into an equation is not new; a well-known earlier instance appears in the \emph{method of manufactured solutions} \cite{mms}, which is widely used for code verification when developing numerical solvers. By constructing an exact solution, one can compare a numerical approximation against this exact solution. The novelty of our work is that we explore the use of this general concept in a completely different setting: for generating training data for neural PDE solvers, in order to obtain solutions to a family of PDE problems without ever solving the PDE. We do this by coupling manufactured solutions with classical PDE theory to randomly draw unknown functions from the solution space, creating a training dataset that generalizes effectively for operator learning. In particular, our experiments demonstrate that models trained solely on data produced by our “backwards” method still achieve strong performance when tested on data generated by conventional numerical solvers, highlighting generalizations offered by our approach. For more details see Section \ref{numerical_experiments}.

Recall the standard supervised learning setting where the training data are input-output pairs $(x_j, y_j),$ where the input vectors $x_j$ are independent and identical draws from an underlying distribution ${\cal D}$, and $y_j = {\cal G}(x_j)$, and the goal is to derive an approximation ${\tilde{{\cal G}}}$ with minimal test error $\mathbb{E}_{x \sim {\cal D}} | \tilde{{\cal G}}(x) - {\cal G}(x) |$.  In our setting, the function to learn is the operator ${\cal G}: (a, f) \rightarrow u$.  Our method of generating $(a_j, f_{a_j}^k)$  and our overall approach can be viewed as a best attempt within the operator learning framework to replicate training data within the classical supervised learning setting.

\textbf{Organization of the paper.}  This paper is organized as follows: in Section \ref{set-up} we introduce the main idea in more detail, in Section \ref{data_gen} we discuss how to determine a space for the unknown functions depending on the problem, and in Section \ref{numerical_experiments} we present numerical experiments using our data in a known network architecture such as the Fourier Neural Operator \cite{NeuralOperator} (FNO). The types of PDE problems we consider are elliptic linear and semi-linear second-order equations with Dirichlet and Neumann boundary conditions, starting with the Poisson equation as a first example and then considering more complicated equations. In Subsection \ref{subsection:comparison_with_classical_solvers}, we present experiments comparing our method to a more classical approach, where the right-hand side is first generated and the corresponding problem is solved numerically to obtain input-output pairs for training. At the end of this paper, we include an appendix section with a description of the mathematical symbols used in this paper. Our data generation code can be found on GitHub under the repository name \href{https://github.com/erisahasani/synthetic-data-for-neural-operators}{synthetic-data-for-neural-operators}.

\section{Set-up and Main Approach}\label{set-up}
Consider a class of PDE problems of the form

\begin{equation} \label{1.1}
\left\{\begin{aligned}
-L_au\, &=f & \text{in } \Omega\\[5pt]
B(u) & = 0 & \text{on } \partial{\Omega} ,
\end{aligned}\right.
\end{equation}
where $L = L_{a}$ denotes a differential operator parameterized by $a \in \mathcal{A}$, where $a$ here denotes some abstract parametrization which depends on the type of PDE, see for example subsection \ref{subsection: linear}. $\Omega \subset \mathbb{R}^n$ is a given bounded domain, and $B(u)$ denotes a given boundary condition. The goal is to find a solution $u$ that solves \eqref{1.1} given $L_a$ and $f$. So in a general setting, we wish to learn an operator of the form 
\begin{align*}
    \mathcal{G}: \mathcal{A}\times \mathcal{F} &\longrightarrow \mathcal{U} \\
    (a,f) &\longmapsto u,
\end{align*}
where $\mathcal{A}, \mathcal{F}$ and $\mathcal{U}$ are function spaces that depend on the specifics of the PDE problem.

For example, if we take $Lu = \Delta u$, and $B(u) = u$ then \eqref{1.1} becomes the Poisson equation with zero Dirichlet boundary condition. In this case, we can take $\mathcal{F} = L^2(\Omega)$ and $\mathcal{U}=H^1_0(\Omega)$ and the operator we wish to learn is of the form
\begin{align*}
    \mathcal{G}: L^2(\Omega) &\longrightarrow H^1_0(\Omega) \\
    f &\longmapsto u,
\end{align*}

So instead of first fixing a function $f$ and then solving \eqref{1.1} to obtain $u$ to be used as input-output pairs $(f,u)$, we instead generate $u$ first, plug it into $\eqref{1.1}$, and compute $f$ by the specified rule. 

The main innovation of our work is in determining the appropriate class of functions for the unknown function $u$ in \eqref{1.1}. While from the PDE theory we know that $u$ lives in some Sobolev space (see e.g \cite{evans2010partial}) in the case of elliptic PDEs, such space is infinite-dimensional and it is unclear at first how to generate functions that serve as good representatives of the full infinite-dimensional space. We propose to generate functions as random linear combinations of basis functions of the corresponding Sobolev space. In the case where we know from theory that the underlying Sobolev space is $H^1_0(\Omega)$ or $H^1(\Omega)$, then we can obtain explicit basis elements that can be obtained by the eigenfunctions of the Laplace operator with Dirichlet and Neumann boundary conditions, respectively.

\section{Drawing synthetic representative functions from a Sobolev space}\label{data_gen}
In this section, we discuss how to draw representative functions from the solution space in the case of elliptic problems so that they generalize well when used in numerical experiments. See the appendix for the definitions of the function spaces used in this section.

Let $\Omega \subset \mathbb{R}^n$ be a bounded open set. Consider the following eigenvalue problem 
\begin{equation} \label{eigenvalue_problem}
\left\{\begin{aligned}
- \Delta u &= \lambda u && \text{in } \Omega\\[5pt]
B(u)&= 0 && \text{on } \partial{\Omega},
\end{aligned}\right.
\end{equation}
which is called the Laplace-Dirichlet operator when $B(u) = u$. We say that $\lambda \in \mathbb{R}$ is an eigenvalue to the Laplace-Dirichlet operator if there exists $u \in H_0^1(\Omega)$ with $u \neq 0$ such that 
\begin{align*}
    \int_{\Omega} \nabla u(x) \nabla \varphi(x) dx = \lambda \int_{\Omega} u(x) \varphi(x) dx, \text{ for all } \varphi \in H_0^1(\Omega)
\end{align*}
If such $u \neq 0$, we say that it is an eigenfunction associated to the eigenvalue $\lambda$. The following theorem is well known in the analysis of PDEs and spectral theory (see Chapter 8 of \cite{attouch2014variational}).

\begin{theorem}\label{eigen_theorem} The Laplace-Dirichlet operator has countably many eigenvalues $0< \lambda_1 \le \lambda_2 \le \cdots \lambda_N \le \cdots$. 
There exists an orthonormal basis $(e_i)_{i=0}^{\infty}$ of $L^2(\Omega)$ such that $e_i$ is an eigenfunction of the Laplace-Dirichlet operator, i.e of problem \eqref{eigenvalue_problem}, corresponding to the eigenvalue $\lambda_i$ for each $i \in \mathbb{N}$. Moreover, $(e_i\!/ \sqrt{\lambda_i})_{i=0}^{\infty}$ is an orthonormal basis of $H_0^1(\Omega)$ equipped with the scalar product $\langle u, \varphi \rangle = \int_{\Omega} \nabla u \cdot \nabla \varphi$.
\end{theorem}

This theory extends to more general Hilbert spaces, including different elliptic linear operators or different types of boundary value conditions such as Neumann or mixed (e.g. see Theorem 6.6.1 in \cite{attouch2014variational}).

For Neumann boundary conditions, we have $B(u) = \partial u \cdot \nu$ where $\nu$ denotes the exterior unit normal vector to the boundary $\partial \Omega$ in problem \eqref{eigenvalue_problem}, then we have a similar theorem for the Hilbert space $V = \{ v \in H^1(\Omega): \int_{\Omega} v(x)dx =0 \}$, where we can obtain an orthogonal basis for the functional space $V$. Notice that $V$ here is essentially $H^1(\Omega)$ but functions that differ by adding or subtracting a constant are considered the same.

\subsection{Representative functions in rectangular domains}
Eigenfunctions of the Laplace operator are known for the Dirichlet, Neumann, and Robin boundary conditions on rectangular domains of the form $(a_1,b_1) \times (a_2,b_2)\times \cdots \times (a_n, b_n) \subset \mathbb{R}^n$. They are also known for some non-rectangular domains, see Figure \ref{fig:triangle_example} for an example on a triangular domain. To keep the presentation simple, we will mainly consider Dirichlet ($B(u) :=0$) and Neumann ($B(u) := \nabla u \cdot \nu$) boundary conditions on $\Omega := (0,1)^2$.

For the Dirichlet case, the eigenfunctions $e_{ij}$ corresponding to the eigenvalues $\lambda_{ij}$ of problem \eqref{eigenvalue_problem} are given by
\begin{equation}\label{eigen_sine}
    e_{ij}(x,y) = \sin(i \pi x) \sin(j \pi y), \quad \lambda_{ij} = (i \pi)^2 + (j \pi)^2, \quad (x,y) \in (0,1)^2, i,j \in \mathbb{N}.
\end{equation}
For the Neumann case, they are given by
\begin{equation}\label{eigen_cosine}
    e_{ij}(x,y) = \cos(i \pi x) \cos(j \pi y), \quad \lambda_{ij} = (i \pi)^2 + (j \pi)^2, \quad (x,y) \in (0,1)^2, i,j \in \mathbb{N}.
\end{equation}
Further, normalizing appropriately, we define the following basis elements for $H_0^1(\Omega)$ and $V$, respectively
\begin{equation}\label{basis_functions}
    u_{ij}(x,y) := \frac{\sin(i \pi x) \sin(j \pi y)}{\sqrt{(i \pi)^2 + (j \pi)^2}}, \quad v_{ij}(x,y) = \frac{\cos(i \pi x) \cos(j \pi y)}{\sqrt{(i \pi)^2 + (j \pi)^2}}
\end{equation}

\begin{remark}\label{remark1}
    Notice that since the $u_{ij}$ are basis elements of $H_0^1(\Omega)$, for any $w \in H_0^1(\Omega)$, there exist coefficients $c_{ij}$ such that $w$ can be written precisely as
    \begin{align*}
        w(x,y) = \sum_{i,j =1}^\infty c_{ij} u_{ij}(x,y)
    \end{align*}
    where $c_{ij} = (w,u_{i,j})_{H_0^1}$.
\end{remark}

Keeping the above remark in mind, we generate the unknown functions $u$ (which we assume are from $H_0^1(\Omega)$ or $V$) as truncated sums of random linear combinations basis functions with prescribed decay in the coefficients. More precisely, let $M, K$ denote positive truncation numbers and let $a_{ij}, b_{ij} \sim N(0,1\!/\sqrt{i^2+j^2})$ generate $u \in H_0^1(\Omega)$ and $v \in V$ as follows
\begin{equation}\label{u}
    u(x,y) = \sum_{i=1}^M \sum_{j=1}^K a_{ij} u_{ij}(x,y), \quad v(x,y) = \sum_{i=1}^M \sum_{j=1}^K b_{ij} v_{ij}(x,y)
\end{equation}

Notice that by construction, functions of the form \eqref{u} satisfy zero Dirichlet and zero Neumann boundary conditions, respectively. In experiments, we draw $M$ and $K$ randomly in $\{ 1, 2, \cdots, 20\}$, that is to say we use up to the first $20$ basis functions. While these spaces are infinite-dimensional and thus require an infinite number of basis functions, we observe that truncating to the first $M=20$ is sufficient to achieve good generalizations to unseen non-trigonometric $f$ functions.

\begin{remark}
    While in the eigenvalue problem \eqref{eigenvalue_problem} we are seeking eigenfunctions and eigenvalues of the Laplace-Dirichlet operator, as we can see in experiments later (see e.g. Section \ref{subsection:nonlinear}), these functions can be used for nonlinear elliptic PDEs as well. This is because Theorem \ref{eigen_theorem} asserts that $(e_i\!/ \sqrt{\lambda_i})_{i=0}^{\infty}$ is an orthonormal basis for $H_0^1(\Omega)$, which means that as long as we know apriori that a solution belongs to $H_0^1(\Omega)$, the same eigenfunctions are `good' representative.
\end{remark}

\subsection{Representative functions in non-rectangular domains}
Our experiments mainly focus on square domains; however, the eigenfunctions and eigenvalues of the Laplacian are also known explicitly for certain specific non-rectangular domains. In particular, the Laplacian eigenfunctions are known for disks, circular annuli, spheres and spherical shells which can generally be described as $\Omega = \{ x \in \mathbb{R}^n: r < |x| < R \}$, with $n=2,3$, as well as for ellipses and elliptical annuli. In addition, they are known for equilateral triangles, that is when $\Omega = \{ (x,y) \in \mathbb{R}^2: 0<x<1, 0<y<\sqrt{3}x, y < \sqrt{3}(1-x) \}$. For more details on eigenfunctions of the Laplacian, see \cite{laplace_eigen}.

As for domains that are not of the above type, there could be ways to obtain the eigenfunctions of the Laplacian numerically; however, in this case, we cannot easily take derivatives symbolically -- the main reason our method is computationally efficient. A potential way to generalize to any domain shape could be by passing the boundary values as an input during the training phase and asking for the right boundary condition after training is finished to get a prediction; this is an interesting future direction to explore.

\section{Numerical Experiments using the Fourier Neural Operator}\label{numerical_experiments}
\label{sec:others}

In this section, we perform several numerical experiments to demonstrate the capabilities of our method. Through some linear and non-linear PDE examples, we show how our method generalizes for $f \in L^2(\Omega)$, see subsections \ref{subsection:poisson} and \ref{subsection:nonlinear}. Namely, after training FNO with our data, we pick an $f$ that is not of trigonometric form, use a numerical solver to obtain a solution $u$ (at high resolution for better accuracy, which is then down-sample to $85 \times 85$) and then see how this ("exact") solution  $u$ compares to the solution predicted by FNO when trained only with our data. In subsection \ref{subsection:comparison_with_classical_solvers} we compare our method to a more classical approach of first generating $f$, then solving numerically for $u$.

The architecture we use for numerical experiments is the Fourier Neural Operator (FNO) introduced in \cite{NeuralOperator}, which can learn mappings between function spaces of infinite-dimensions. The advantage of FNO is that it aims to approximate an operator that learns to solve a family of PDEs by mapping known parameters to the solution of that PDE, instead of only approximating one instance of a PDE problem. Due to the nature of FNO, this enables us to use our synthetic data in order to approximate an entire class of problems at once. The novelty of FNO is that the kernel function, which is learned from the data, is parameterized directly in Fourier space, leveraging the Fast Fourier Transform when computing the kernel function. We train FNO using Adam optimizer on batches of size $100$, with a learning rate of $0.001$, modes set to $12$, and of width $64$. We also use relative $L_2$ error to measure performance for both training and testing.

We focus on second-order semi-linear elliptic PDE equations in divergence form defined on $\Omega:= (0,1)^2$, with zero boundary conditions, given by 
\begin{equation} \label{second_order_elliptic}
\left\{\begin{aligned}
-\operatorname{div}(A(x) \cdot \nabla u) + b\cdot \nabla u + cu+ g(u) &=f && \text{in } \Omega\\[5pt]
B(u) & = 0 && \text{on } \partial{\Omega},
\end{aligned}\right.
\end{equation}
 where $A(x) \in \mathbb{R}^{2 \times 2}$, $b \in \mathbb{R}^2$, $c \in \mathbb{R}$ and $g(u)$ is some nonlinear function in $u$ and $B(u)$ is either $B(u)=u$ or $B(u) = \nabla u \cdot \nu$, where $\nu$ is the exterior unit normal vector to the boundary $\partial \Omega$ that correspond to zero Dirichlet or Neumann condition, respectively. Here we also assume that $A$ is uniformly elliptic and each $a_{ij} \in L^{\infty}(\Omega)$ with $i,j \in \{1,2 \}$.

For the rest of the paper, we will denote by $H(\Omega)$ the corresponding Sobolev space depending on $B(u)$, which is $H(\Omega) = H^1_0(\Omega)$ when $B(u)=u$ and $H(\Omega) = H^1(\Omega)$ when $B(u) = \nabla u \cdot \nu$.

\subsection{The Poisson Equation}\label{subsection:poisson}
We first consider a simple example of problem \eqref{second_order_elliptic}, the Poisson equation, by taking $A(x) = I$  (the $2\times 2$ identity matrix), $b=(0,0)$, $c=0$ and $g(u) \equiv 0$
\begin{equation} \label{Poisson}
\left\{\begin{aligned}
-\Delta u\, &=f && \text{in } \Omega\\[5pt]
B(u) & = 0 && \text{on } \partial{\Omega},
\end{aligned}\right.
\end{equation}
Our goal is to learn an operator of the form:
\begin{align*}
    G: L^2(\Omega) & \longrightarrow H(\Omega) \\
    f & \longmapsto u
\end{align*}

Notice that the Poisson equation can be easily solved when fixing $f$, however, here we would like to demonstrate our method of generating data on this easy problem first. Later we will consider more complicated examples. 

We generate data points of the form $(f,u)$ where $u$ is defined as in \eqref{u}, depending on $B(u)$, and $f$ is computed by taking derivatives of $u$ so that \eqref{Poisson} holds. This way, we can generate a lot of data. We let $M$ and $K$ in \eqref{u} range between $1$ and $20$, so that we can get a variety of such functions and various oscillations. We perform experiments by training with $1000$, $10000$, and $100000$ functional data points and testing with $100$ data points for the Poisson problem with Dirichlet and then with Neumann boundary conditions. We report the relative $L_2$ errors in the following Table \ref{table1}. 

\begin{table}[h!]
\begin{center}
\begin{tabular}{p{2.5cm}p{5cm}p{5cm}}
 \ & \textbf{Dirichlet }& \textbf{Neumann }\\
\end{tabular}
\begin{tabular}{|p{2.5cm}||p{1.5cm}|p{1.5cm}|p{1.5cm}|p{1.5cm}|p{1.5cm}|p{1.5cm}|}
\hline
\textbf{Training points} & 1,000 &10,000 &100,000& 1,000 &10,000 &100,000 \\[0.5pt]
\hline\hline
\textbf{Training loss} & 0.01359  & 0.00322&  0.00078& 0.00818  & 0.00232 & 0.00065 \\
 \hline
\textbf{Testing loss} & 0.02266 & 0.00346 & 0.00072 & 0.01877 & 0.00298 & 0.00066 \\ 
\hline
\end{tabular}
\caption{\label{table1}FNO performance on the Poisson equation using our synthetic data generated as in \eqref{u}.}
\end{center}
\end{table}

\textbf{Testing on $f$ beyond finite trigonometric sums}. Notice that if $u$ is represented as a finite linear sum of sines and cosines, as in \eqref{u}, then $f$ generated according to \eqref{Poisson} also consists of a finite linear sum of sines or cosines depending on $B$. So it is important to test on $f$'s that are not sums of sines or cosines to demonstrate that our method of generating data generalizes well.

Restricting our attention to the Dirichlet case, let us generate $f$ so that it does not consist of sine or cosine functions.  This is akin to out-of-distribution testing in the machine learning literature.  We consider the following two example functions: $f_1(x,y) = x-y$, which is smooth, and $f_2(x,y) = |x-0.5||y-0.5|$, which is a not everywhere differentiable function. However, in each case, $f_1, f_2$ are in $L^2(\Omega)$, and approximation of $L^2$ functions by trigonometric functions is well studied, and error bounds are available (see \cite{devore1993constructive}). So we expect to obtain approximate solutions to the Poisson equation \eqref{Poisson} for any $f$ function that is in $ L^2(\Omega)$.

\begin{figure}[!ht]

\begin{center}
\centerline{\includegraphics[width=\columnwidth]{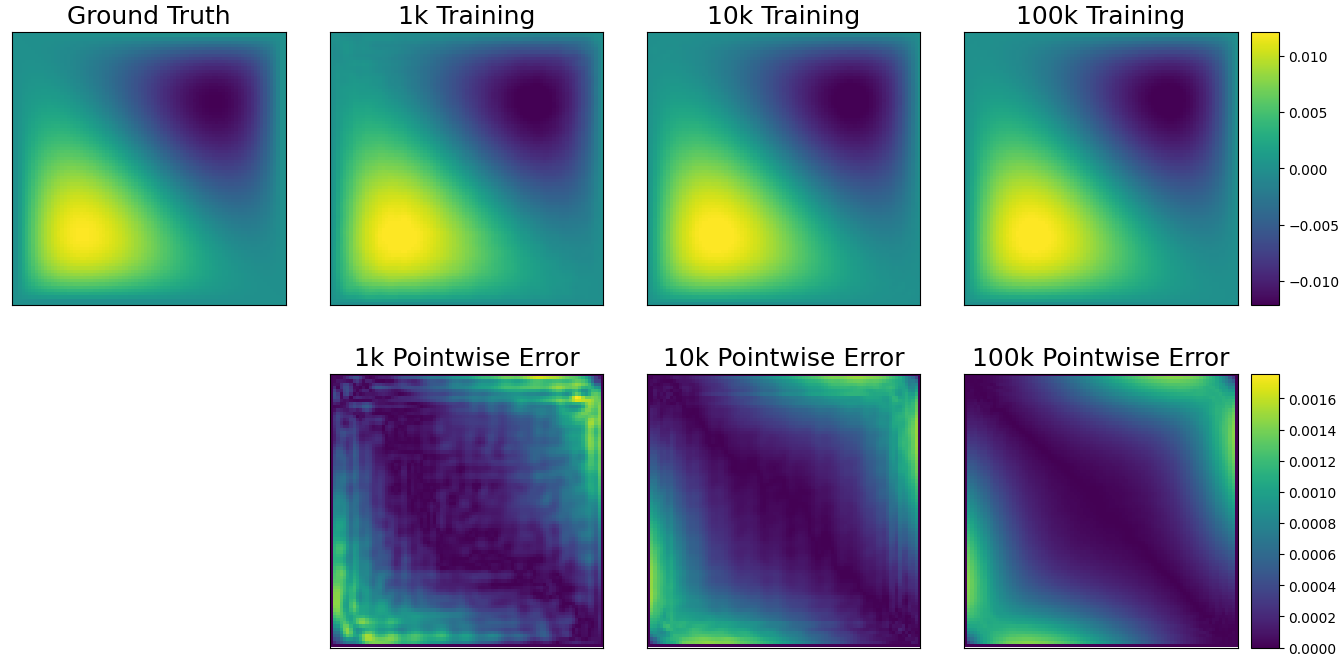}}
\caption{ Predicted solutions of the Poisson equation \eqref{Poisson} with $f_1(x,y)=x-y$ as the right-hand side using FNO with $1,000$, $10,000$ and $100,000$ training data points. Their relative $L^2$ errors are $0.097$, $0.096$ and $0.094$, respectively, while their relative $l^\infty$ errors are $0.145, 0.123$ and $0.118$, respectively.  }
\label{fig1a}
\end{center}

\end{figure}

\begin{figure}[!ht]

\begin{center}
\centering{\includegraphics[width=\columnwidth]{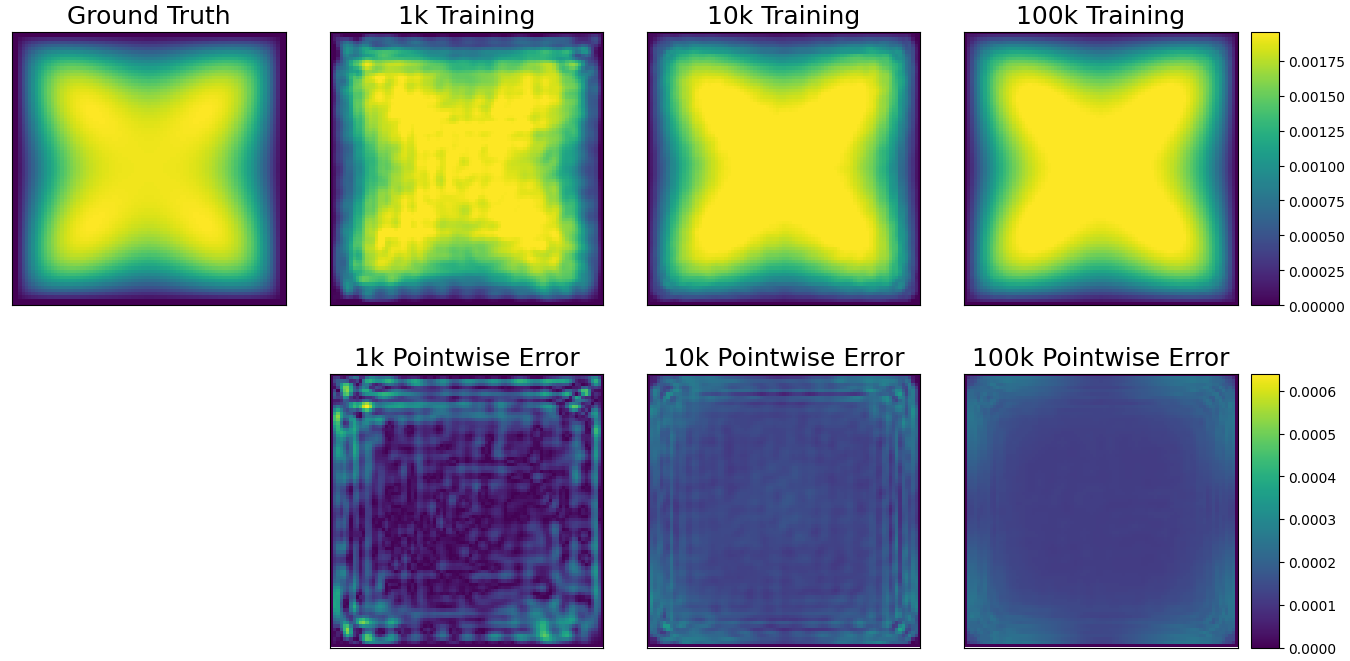}}
\caption{ Predicted solutions of the Poisson equation \eqref{Poisson} with $f_2(x,y) = |x-0.5||y-0.5|$ as the right-hand side using FNO with $1,000$, $10,000$ and $100,000$ training data points. Their relative $L^2$ errors are $0.102$, $0.114$ and $0.108$, respectively, while their relative $l^\infty$ errors are $0.326, 0.188$ and $0.131$, respectively. }
\label{fig1b}
\end{center}

\end{figure}

In Figures \ref{fig1a} and \ref{fig1b}, we summarize the predicted solutions using FNO, when trained with $1,000$, $10,000$ and $100,000$ synthetic data functions that consist of sine functions given by \eqref{u}. We also record the relative $L^2$ errors for each example. The following demonstrates that the choice of functions constructed as in \eqref{u} generalizes well. 

Note that FNO performs better when predicting a solution to the Poisson equation when the right-hand side is given by a smooth function, and has a harder time when the right-hand side is not smooth in $\Omega$. This behavior is expected from classical Fourier analysis. Smooth functions can be well-approximated by trigonometric polynomials, with the error decaying like $O(1\!/n^k)$ if the function is $C^k$, where $n$ denotes the truncation number and $k$ denotes the number of continuous derivatives. For non-smooth functions, although pointwise convergence still holds, uniform convergence may fail and the approximation error decays much more slowly, explaining the drop in FNO performance from Figure \ref{fig1a} to Figure \ref{fig1b}.

\subsection{Second-order semi-linear elliptic PDE}\label{subsection:nonlinear}
We take $A = I$, $b=(0,0)$, $c=0$ and $g(u) = u^2$ in \eqref{second_order_elliptic}, in which case the problem becomes 
\begin{equation}\label{semilinear}
\left\{\begin{aligned}
-\Delta u + u^2 &=f && \text{in } \Omega\\[5pt]
B(u) & = 0 && \text{on } \partial{\Omega},
\end{aligned}\right.
\end{equation}
In this problem we have a nonlinear term $g(u)=u^2$ added. It turns out that despite the nonlinear term, we get decent approximations of solutions when using our data generation with FNO.

As before, we generate $u$ as specified in \eqref{u} and compute $f$ by plugging into \eqref{semilinear}. From the theory, we know that the space of solutions is $H(\Omega)$. Numerical experiments show that despite the non-linearity in that term, FNO achieves low $L_2$ relative errors, as indicated in Table \ref{table4}.  We summarize the relative $L_2$ errors of training and testing loss in Table \ref{table4} when we train on $1,000$, $10,000$ and $100,000$ data points and test on $100$ data points.

\begin{table}[h!]
\begin{center}
\begin{tabular}{p{2.5cm}p{5cm}p{5cm}}
 \ & \textbf{Dirichlet }& \textbf{Neumann }\\
\end{tabular}
\begin{tabular}{|p{2.5cm}||p{1.5cm}|p{1.5cm}|p{1.5cm}|p{1.5cm}|p{1.5cm}|p{1.5cm}|}
\hline
\textbf{Training points} & 1,000 &10,000 &100,000 & 1,000 &10,000 &100,000 \\[0.5pt]
\hline\hline
\textbf{Training loss} & 0.01679 & 0.01763  & 0.00184 & 0.01017 & 0.00800  & 0.00237 \\
 \hline
\textbf{Testing loss} & 0.03391 & 0.02693 & 0.00562 & 0.02992 & 0.01472& 0.00295 \\ 
\hline
\end{tabular}
\caption{\label{table4}FNO performance on the problem \eqref{semilinear} using \eqref{u} functions.}
\end{center}
\end{table}

\textbf{Testing on $f$ beyond finite trigonometric sums.} 
When we generate the unknown $u$ to be of sums of sines or cosines, when plugging in equation \eqref{semilinear}, the computed f still consists of sines and cosines, but with some terms squared. As before, after only training FNO with such $u$'s, we are interested in seeing how well it generalizes when testing non-trigonometric $L^2(\Omega)$ functions. We demonstrate generalization through the following two examples: $f_1(x,y) = xy$ and $f_2(x,y) = (x-0.5)^2 + (y-0.5)^2$.

\begin{figure}[!ht]

\begin{center}
\centerline{\includegraphics[width=\columnwidth]{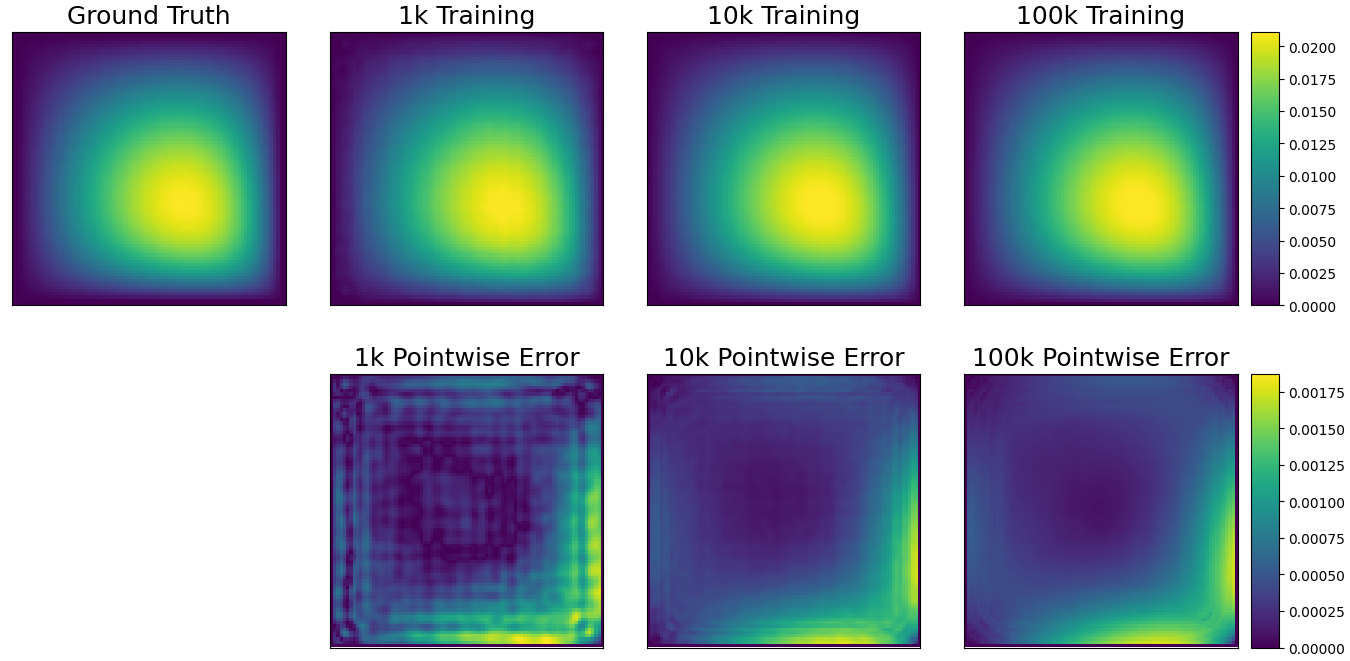}}
\caption{  Predicted solutions of the semi-linear equation \eqref{semilinear} with $f_1(x,y)=xy$ as the right-hand side using FNO with $1,000$, $10,000$ and $100,000$ training data points. Their relative $L^2$ errors are $0.058$, $0.058$ and $0.059$, respectively, while their relative $l^\infty$ errors are $0.089, 0.084$ and $0.083$, respectively. }
\label{fig2a}
\end{center}

\end{figure}

\begin{figure}[!ht]

\begin{center}
\centering{\includegraphics[width=\columnwidth]{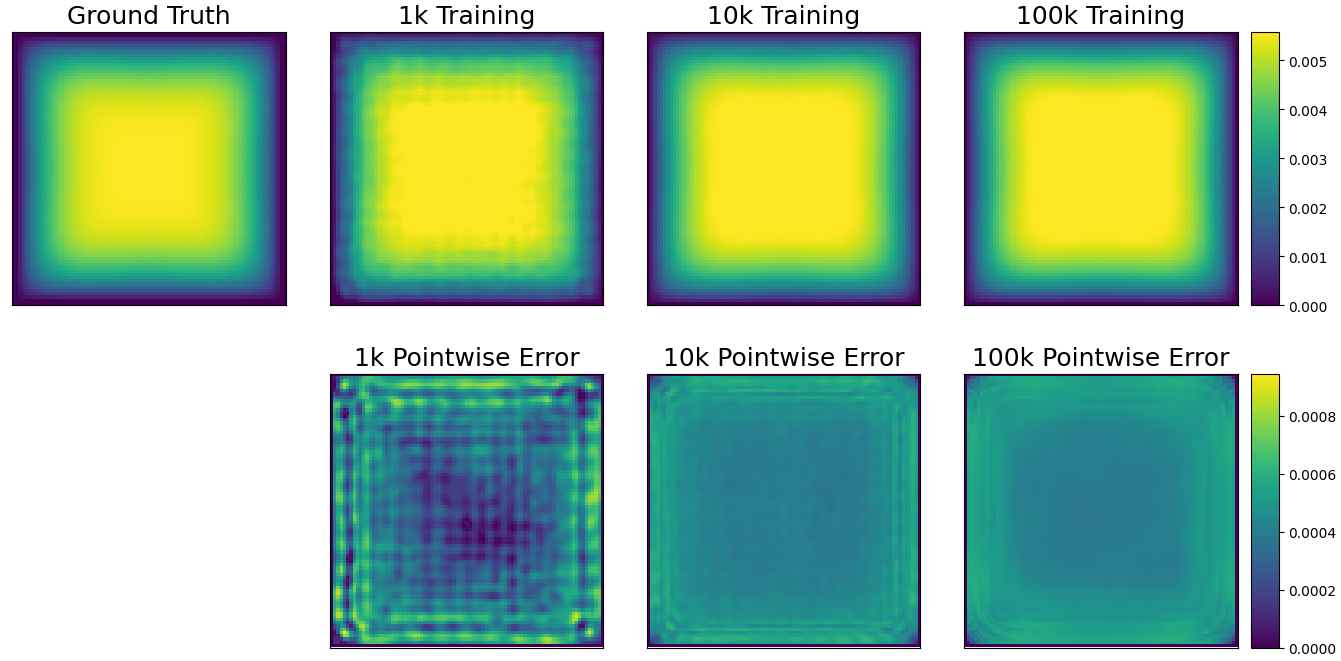}}
\caption{  Predicted solutions of the semi-linear equation \eqref{semilinear} with $f_2(x,y) = (x-0.5)^2 + (y-0.5)^2$ as the right-hand side using FNO with $1,000$, $10,000$ and $100,000$ training data points. Their relative $L^2$ errors are $0.107$, $0.115$ and $0.118$, respectively, while their relative $l^\infty$ errors are $0.326, 0.113$ and $0.106$, respectively. }
\label{fig2b}
\end{center}
\end{figure}

The error plateaus after a certain amount of training data and stops decreasing further, even though the predicted solution becomes  smoother. We notice similar behavior across several right-hand sides in equation \eqref{semilinear} that smooth, albeit not finite sums of sines or cosines.

\subsection{Comparison to a more classical approach}\label{subsection:comparison_with_classical_solvers}

In this subsection, we compare our backward generation method for producing training data with a more classical forward approach based on solving the PDE numerically for randomly generated right-hand sides. Consider the same example as above, equation \eqref{semilinear}. We compare our method of generating data to using classical numerical solvers instead. In the forward generation method, we first draw $f \in L^2(\Omega)$ and then use a numerical solver to obtain a solution for the corresponding $u$. 

We highlight a few difficulties with this approach: it is computationally more expensive to solve the PDE for each new $f$; it introduces additional error into the training data due to reliance on approximate numerical solutions; and in problems involving nonlinearity, like in example \eqref{semilinear}, uniqueness of solutions is not guaranteed.

To the best of our effort to provide a fair comparison, we draw $f$'s from $L^2(\Omega)$ by considering truncated series of orthonormal eigenvectors of $L^2(\Omega)$ with random coefficients. Recall the first part of Theorem \ref{eigen_theorem}, where eigenfunctions of the Laplace-Dirichlet operator form an orthonormal basis for $L^2(\Omega)$. Since we aim to solve \eqref{semilinear} for every $f \in L^2(\Omega)$, it is natural to generate $f$'s as truncated linear combinations of such basis elements. Specifically, for any $f \in L^2(\Omega)$, we can write
\begin{align*}
    f(x,y) = \sum_{i,j =1}^{\infty} \langle f,e_{i,j} \rangle e_{i,j},
\end{align*}
with $e_{i,j}$ defined as in \eqref{eigen_sine} or \eqref{eigen_cosine}. 

Intuitively, we are comparing two strategies: (1) approximating the $H_0^1$ space by generating representative $u$'s and directly computing the corresponding $f$ (our backward method), versus (2) approximating the $L^2$ space by drawing representative $f$'s and solving the PDE numerically to obtain $u$ (the forward method), which introduces additional numerical errors.

Our experiments indicate that our backward method is not only faster but also provides better accuracy when tested on non-trigonometric $f$'s. For the experiments, both training and testing data have resolution $85 \times 85$, and we maintain the same resolution when generating training data via truncated series of basis functions in the forward method. It is important to note that discretization introduces error into the training set, and reducing this error requires increasing the grid resolution, which significantly slows down data generation. By contrast, our backward method does not introduce discretization error because the solutions are computed symbolically and exactly.

We trained the Fourier Neural Operator (FNO) model with $10,000$ training samples generated by each method. Data generation using our backward method took approximately $12$ minutes, leveraging the SymEngine library for symbolic differentiation. In contrast, data generation using the forward method required about $20$ minutes, relying on the FiPy Python library to numerically solve each instance on grids of size $85 \times 85$.

\begin{figure}[!ht]
\begin{center}
\centering{\includegraphics[width=13cm]{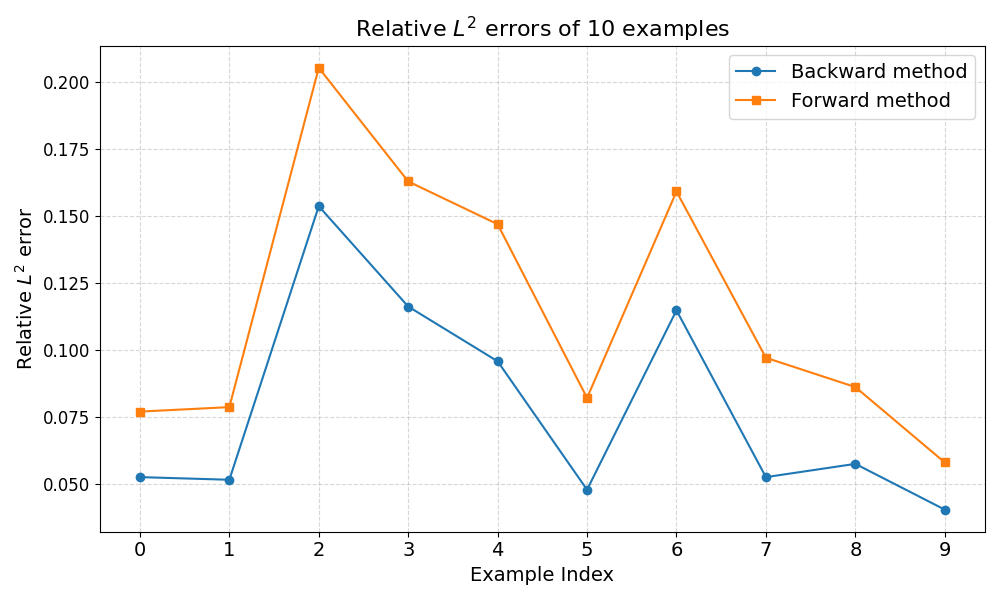}}
\caption{Comparison of our backward generation method with the classical forward generation method, as described in Subsection \ref{subsection:comparison_with_classical_solvers}, on the semi-linear problem \eqref{semilinear}. Relative $L^2$ errors of predicted solutions are shown: blue corresponds to our backward method and orange corresponds to the forward method. Our method consistently achieves lower errors across all test cases.}
\label{fig:forward_vs_backward_method}
\end{center}
\end{figure}

After training on each dataset separately, we generate a set of $10$ functions $f$ used for testing. In order to obtain solutions as precise as possible, we solve for each $f$ on a $1700 \times 1700$ grid and then downsample the results to $85 \times 85$. The $f$'s are given by:
\begin{enumerate}
    \setcounter{enumi}{-1}
    \item $2x(1-x)+2y(1-y)$
    \item $xy$
    \item $1$
    \item $(x-0.5)^2+(y-1)^2$
    \item Step function: $1$ if $x>0.5$, $0$ otherwise
    \item $x-y$
    \item $|x-0.5||y-0.5|$
    \item $(x-0.5)^2-(y-1)^2$
    \item $\frac{1}{1+x^2+y^2}$
    \item $e^{-\sqrt{x^2+y^2}}$
\end{enumerate}
We summarize our findings in Figure \ref{fig:forward_vs_backward_method}.

\subsection{Second-order linear elliptic PDE}\label{subsection: linear}
In problem \eqref{second_order_elliptic}, take $g(u)=0$ and allow the matrix $A$ and the lower-order terms to be of any form, possibly depending on $(x,y)$. Then \eqref{second_order_elliptic} becomes
\begin{equation} \label{second_order_only}
\left\{\begin{aligned}
-\operatorname{div}(A \cdot \nabla u) + b \cdot \nabla u + cu &=f && \text{in } \Omega\\[5pt]
B(u) & = 0 && \text{on } \partial{\Omega},
\end{aligned}\right.
\end{equation}
In general, since we use derivatives in our computations, we assume that the entries of $A$ are once differentiable in the corresponding variables.

\textbf{$ \boldsymbol{A}$ as a fixed matrix}. First, we look at the case where we fix a matrix $A$. Then we compute the derivatives involved for the components of $A$ and save those as well. We generate a function $u$ according to \eqref{u}, plug it in to \eqref{second_order_only}, and then compute $f$. As before, the goal is to learn the operator
\begin{align*}
    G: L^2(\Omega) & \longrightarrow H^1(\Omega) \\
    f & \longmapsto u
\end{align*}
For a numerical experiment, let $A$ be as follows
\begin{equation} \label{matrixA}
    A = \begin{pmatrix}
        x^2 & \sin(xy) \\
        x+y & y
    \end{pmatrix}
\end{equation}

In this case, FNO is learning a family of solutions for a fixed $A$ defined above of the problem \eqref{second_order_only} for varying pairs of $f$ and $u$ functions. This choice of $A$ is not particularly special, and the same process can be repeated for any positive definite $A$ (so that \eqref{second_order_only} is elliptic).  For the most accurate results, we can re-generate data points of the form $(f,u)$ for each new matrix $A$ and train different $A$-dependent neural networks.  The following Table \ref{table2} summarizes the relative $L_2$ errors when using FNO to solve \eqref{second_order_only} when $A$ is given by \eqref{matrixA} and when training with $1,000$, $10,000$ and $100,000$ data points and testing with $100$ data points. 

\begin{table}[h!]
\begin{center}
\begin{tabular}{p{2.5cm}p{5cm}p{5cm}}
 \ & \textbf{Dirichlet }& \textbf{Neumann }\\
\end{tabular}
\begin{tabular}{|p{2.5cm}||p{1.5cm}|p{1.5cm}|p{1.5cm}|p{1.5cm}|p{1.5cm}|p{1.5cm}|}
\hline
\textbf{Training points} & 1,000 &10,000 &100,000 & 1,000 &10,000 &100,000 \\[0.5pt]
\hline\hline
\textbf{Training loss} & 0.01992 & 0.01147 & 0.00262 & 0.03452 & 0.00621 &  0.00229 \\
 \hline
\textbf{Testing loss} & 0.04780& 0.01523 & 0.00274 & 0.08926 & 0.00848 & 0.00215 \\ 
\hline
\end{tabular}
\caption{\label{table2}FNO performance on the problem \eqref{second_order_only} with $A$ given by \eqref{matrixA}, using \eqref{u} functions.}
\end{center}
\end{table}

\textbf{$ \boldsymbol{A}$ as a parametric matrix}. As a more general-purpose approach to solving elliptic PDEs using FNO and synthetic data, we can also attempt to train a single neural network for an entire parameterized family of matrices $A$, by passing $A$ as an input in the training data pair. 
That is, instead of fixing the matrix $A$ in our synthetic data, we vary $A$ within a parameterized class and pass it as input data together with $f$. In other words, the learning operator is of the form $G^\dagger: (f,A) \mapsto u$. For simplicity, we assume here that $A$ is a diagonal matrix of the form
\begin{align*}
    A(x,y) = \begin{pmatrix} \alpha(x,y) & 0 \\
   0 & \delta(x,y)
    \end{pmatrix}
\end{align*}
Here, we vary $\alpha(x,y)$ and $\delta(x,y)$. In other words, the operator we are trying to learn is given by
\begin{align*}
    G: L^2(\Omega) \times L^{\infty}(\Omega) \times L^{\infty}(\Omega)& \longrightarrow H(\Omega) \\
    (f, \alpha, \delta) & \longmapsto u
\end{align*}
To further simplify, we assume the components of $A$ are linear functions in $x,y$, that is
\begin{align*}
    A(x,y) = \begin{pmatrix} m_1x +m_2y & 0 \\
   0 & m_3x + m_4y
    \end{pmatrix}
\end{align*}
where $m_i$'s are uniformly distributed in $[0.1,5]$ and $u$ is generated according to \eqref{u} with $M \in \{1,2,\dots, 10\}$. For each generated data point, we generate a matrix of the above form and a function $u$ according to \eqref{u}, then plug them both in equation \eqref{second_order_only} to compute $f$. Finally, the input data forms a triple $(f, \alpha, \delta)$, while the target is to predict $u$. This way, FNO learns how to solve a \emph{family} of functions satisfying \eqref{second_order_only}. We summarize the relative $L_2$ errors in Table \ref{table3} using FNO when training with $1,000$, $5,000$ and $10,000$ data points. As we can see below and as expected, the performance of the FNO with more degrees of freedom in the input data is worse compared to the case where the matrix $A$ is considered fixed and held constant across all the input data.

\begin{table}[h!]
\begin{center}
\begin{tabular}{p{2.5cm}p{5cm}p{5cm}}
 \ & \textbf{Dirichlet }& \textbf{Neumann }\\
\end{tabular}
\begin{tabular}{|p{2.5cm}||p{1.5cm}|p{1.5cm}|p{1.5cm}|p{1.5cm}|p{1.5cm}|p{1.5cm}|}
\hline
\textbf{Training points} & 1,000 &5,000 &10,000 & 1,000 &5,000 &10,000 \\[0.5pt]
\hline\hline
\textbf{Training loss} & 0.12266 & 0.07352 & 0.03134 & 0.14295 & 0.04639 &  0.01107 \\
 \hline
\textbf{Testing loss} & 0.27257 & 0.10641 & 0.04885 & 0.25906 & 0.07611 &  0.05508\\ 
\hline
\end{tabular}
\caption{\label{table3}FNO performance on the problem \eqref{second_order_only} with varying matrix $A$, using \eqref{u} functions.}
\end{center}
\end{table}

\begin{figure}[h!]
    \centering
    \includegraphics[width=9cm]{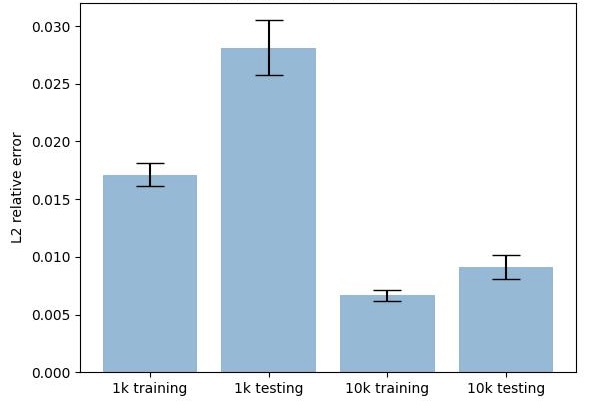}
    \caption{\label{errorbars}Relative $L_2$ errors with standard errors, over 10 experiments with fixed diagonal matrices linear in $x$ and $y$.}
\end{figure}

\subsection{Further examples}\label{subsetion: further examples}

\textbf{Second-order linear elliptic.}
We show an example of a linear second-order where we also include lower-order terms. For example take $A=I$, $b = (3 ,4 )$, $c =1$ and $g(u) =0$. Then \eqref{second_order_elliptic} becomes
\begin{equation} \label{lower_order_terms}
\left\{\begin{aligned}
-\Delta u + 3u_x + 4u_y + u &=f && \text{in } \Omega\\[5pt]
B(u) & = 0 && \text{on } \partial{\Omega},
\end{aligned}\right.
\end{equation}
Once again, we would like to learn the operator
\begin{align*}
G: L^2(\Omega) & \longrightarrow H^1(\Omega) \\
f & \longmapsto u
\end{align*}

\begin{figure}[!ht]

\begin{center}
\centering{\includegraphics[width=\columnwidth]{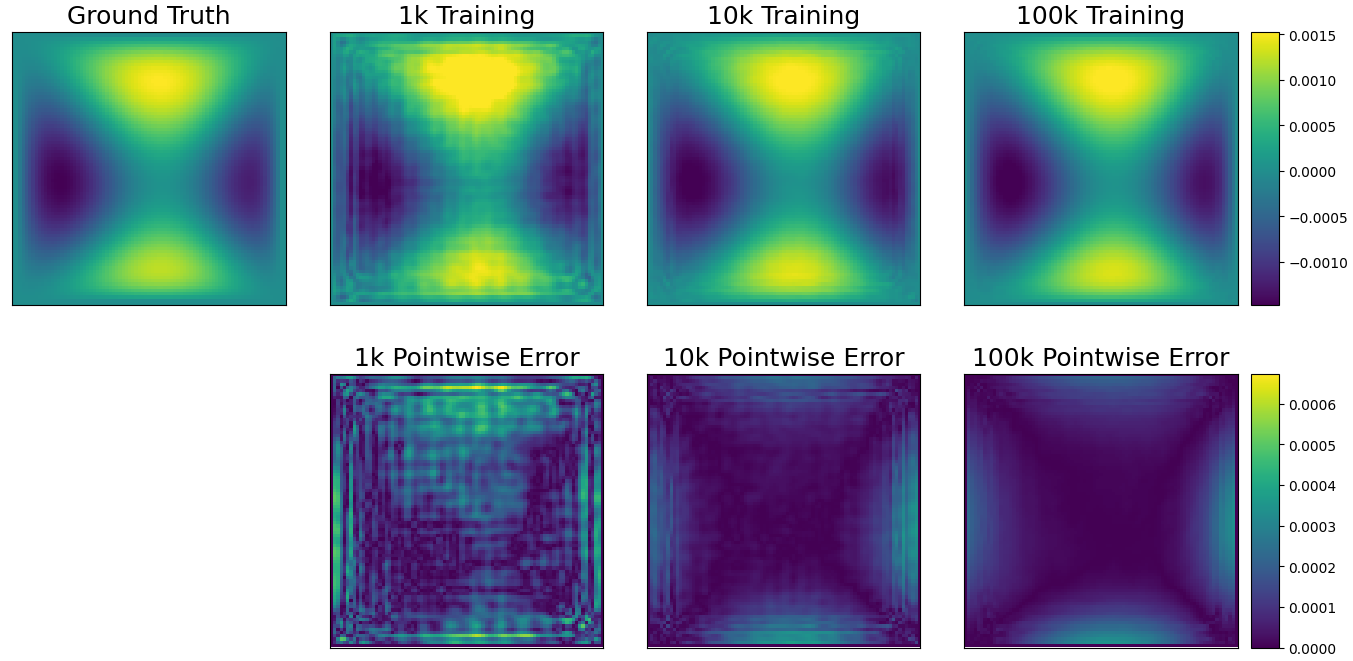}}
\caption{Predicted solutions of the linear equation \eqref{lower_order_terms} with $f(x,y) = (x-0.5)^2 - (y-0.5)^2$ as the right-hand side using FNO with 1,000, 10,000 and 100,000 training data points. Their relative $L^2$ errors are $0.287$, $0.156$ and $0.141$, respectively,  while their relative $l^\infty$ errors are $0.440, 0.251$ and $0.225$, respectively.}
\label{figa1}
\end{center}
\end{figure}

\textbf{Additional semi-linear examples}
Here, we consider problem \eqref{second_order_elliptic} with $A=I$, $b=(0,0)$, $c=0$, $g(u) = \varepsilon e^u$ and $B(u)=u$, that is

\begin{equation} \label{epsilon_problem}
\left\{\begin{aligned}
-\Delta u + \varepsilon e^u &=f && \text{in } \Omega\\[5pt]
u & = 0 && \text{on } \partial{\Omega},
\end{aligned}\right.
\end{equation}

Notice that when $\varepsilon =0$, \eqref{epsilon_problem} becomes the Poisson equation. In this experiment, we let $\varepsilon \in [0,1]$ take values in increments of $0.1$ starting from $0$. The purpose of this is to demonstrate the performance of our method as we go from a linear to a more non-linear problem by recalling the nonlinear term in \eqref{epsilon_problem}. 

For each $\varepsilon =0, 0.1, \cdots ,0.9 ,1.0 $ we generate $10k$ training data and test on $100$, from which five examples are testing data where the right-hand side is first picked and we use a numerical solver to get the solution so we can test on whether we have good generalizations. We record the relative $L^2$ errors of the predicted solution to problem \eqref{epsilon_problem} by fixing the right-hand side $f$ and a $\varepsilon =0, 0.1, \cdots ,0.9 ,1.0$. We summarize the testing performance in the following plot and see that as we go from linear to non-linear, the performance improves, which at first seems surprising.

\begin{figure}[!ht]

\begin{center}
\centering{\includegraphics[width=\columnwidth]{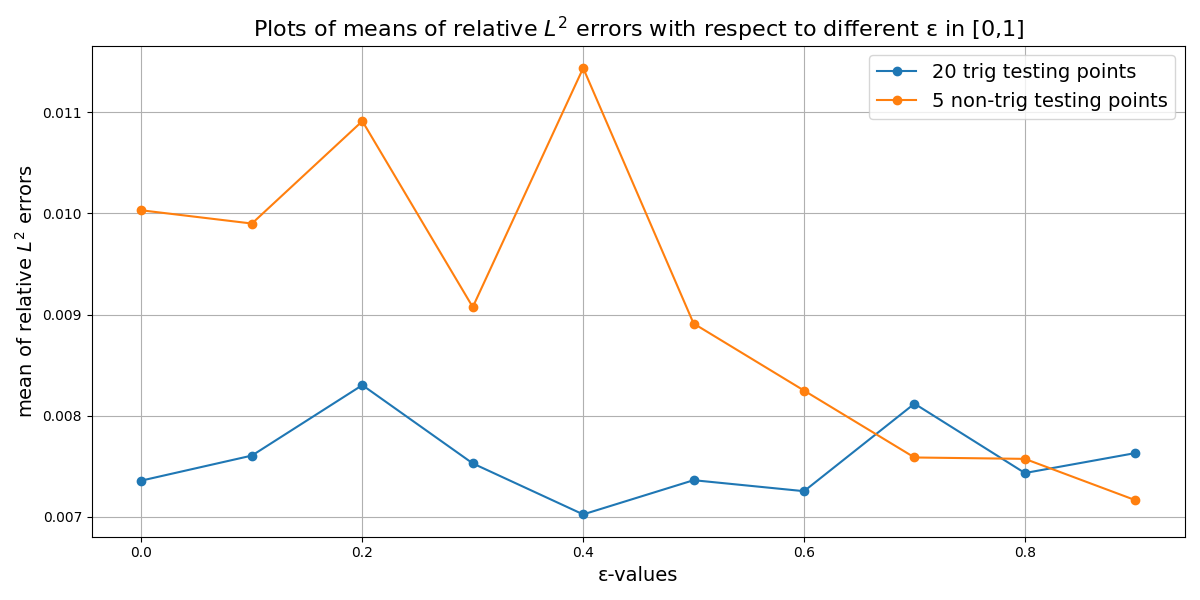}}
\caption{Plots of means of relative $L^2$ errors of 20 testing data for problem \eqref{epsilon_problem} generated using our method and 5 testing data with non-trig right-hand sides for which we invoked numerical solvers to obtain $u$. The five examples of non-trig $f$'s used on the above plot are given by: $x-y, xy, x^2+y-xy,2^{xy},20x-10y. $}
\label{figa2}
\end{center}
\end{figure}

\textbf{The Poisson equation on a triangular domain.}\label{triangle}
Eigenvalues and eigenfunctions of the Laplacian are also known in equilateral triangular domains given by $\Omega = \{ (x,y) \in \mathbb{R}^2: 0<x<1, 0<y<\sqrt{3}x, y< \sqrt{3}(1-x) \}$, first discovered by Lam\'e \cite{triangle_eigen} using reflection and symmetry arguments. We use the first $10$ eigenvalues (some are with multiplicity two) and their corresponding eigenfunctions in order to generate training data. 

\begin{figure}[!ht]
\begin{center}
\centering{\includegraphics[width=\columnwidth]{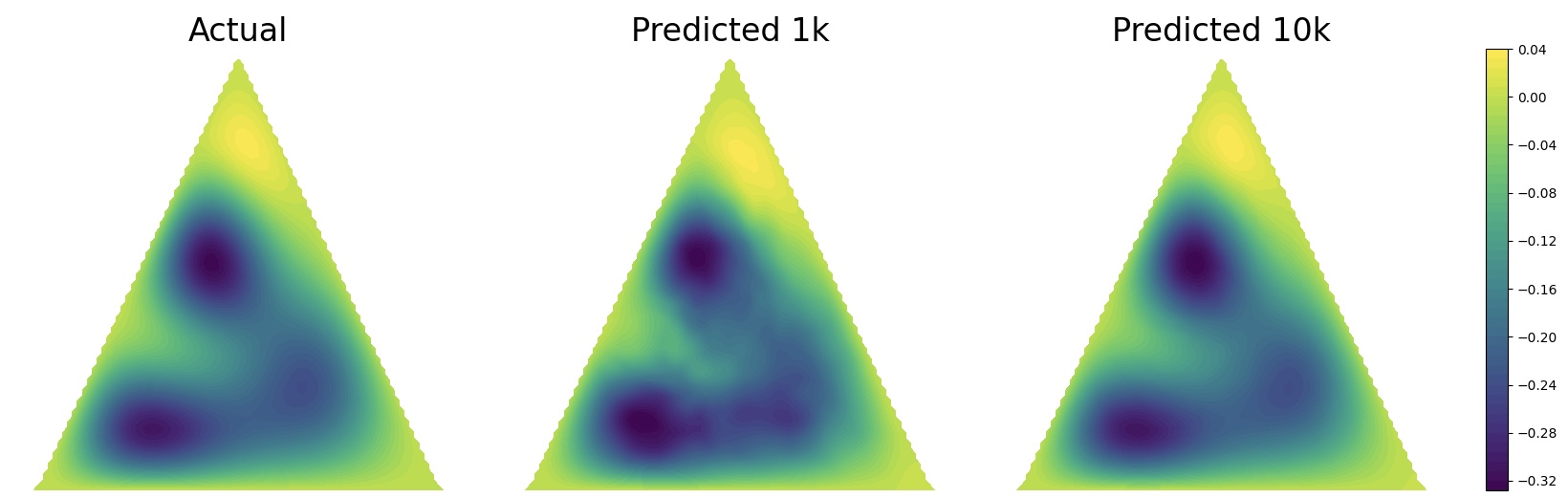}}
\caption{Predicted solutions of an example on a triangular domain using FNO with $1,000$ and $10,000$ training data. The relative $L^2$ errors are respectively $0.10233$ and $0.01424$.}
\label{fig:triangle_example}
\end{center}
\end{figure}

\section{Limitations, conclusion and future work} 

\textbf{Limitations.} Through experiments, we have observed that for certain ``smooth" problems, our method generalizes well. However, as shown in the Darcy flow example (see Section \ref{darcy_example}), where the coefficients are non-smooth, our method did not generalize as effectively. While second-order elliptic PDEs represent a sizable class of problems, there are many other types of PDEs that fall outside this class, such as parabolic and hyperbolic. We have not yet tested the performance of our method on these other types of PDEs. We believe our method could apply to these other cases, but this requires further investigation. Finally, it is worth noting that the selected basis functions are not the only option, and different basis functions may be more suitable for certain problems. Alternative basis functions are worth investigating further.

\textbf{Conclusion and future work.} Using deep learning to solve PDEs has been very promising in recent years. Here, we propose a method that in some settings could eliminate the need to repeatedly solve a PDE for obtaining training data used in training neural operators by first generating the unknown solution and then computing the right-hand side of the equation. Although we exclusively provide theoretical motivation and numerical experiments for second-order elliptic PDEs, this concept could be extended to other types of PDEs where the solution space is known beforehand, enabling the construction of representative functions for such solution spaces. This method could open up the possibility of obtaining good predictions for PDE problems using data-driven neural operators, for which the training data does not require classical numerical solvers to generate.  We stress that our synthetic data generation approach is computationally efficient, particularly compared to solving new PDE problems numerically to generate training data for each new problem instance. We believe that our approach is an important step towards reaching the ultimate goal of \emph{using deep learning to solve PDEs that are intractable using classical numerical solvers}. We also note that as a by-product, our method eliminates sources of error coming from numerically solving PDE problems; instead, our synthetic training data is of the form of exact solutions to a problem on a fixed-size grid. As future work, a promising direction is to use our synthetic data for pre-training, followed by fine-tuning on a few task-specific examples, in line with recent transfer learning approaches explored for elliptic problems (e.g. \cite{herde2024poseidon} and \cite{subramanian2023towards}).

\section*{Acknowledgments}

EH and RW were supported in part by AFOSR MURI FA9550-19-1-0005, NSF DMS-1952735, NSF IFML grant 2019844, NSF DMS-N2109155, and NSF 2217033 and NSF 2217069.

\bibliographystyle{abbrv}

\newpage
\section{Appendix}
\subsection{Notation}
Notation and descriptions used in this paper.
\begin{table}[htbp]
\begin{center}
\begin{tabular}{r c p{10cm} }
\multicolumn{3}{c}{\textbf{Function Spaces}}\\
\toprule
$L^2(\Omega)$ & \quad &  space of Lebesgue-measurable functions $u: \Omega \to \mathbb{R}$ with finite norm $||u||_{L^2} = \left( \int_{\Omega} |u|^2 dx\right)^{1\!/2}$.  \\[5pt]
$L^\infty(\Omega)$ & \quad &  space of Lebesgue-measurable functions $u: \Omega \to \mathbb{R}$ that are essentially bounded.  \\[5pt]
$H^1(\Omega)$ & \quad & Sobolev space of functions $u \in L^2(\Omega)$ with $|\nabla u| \in L^2(\Omega)$, equipped with the inner product $\langle u,v \rangle = \int_{\Omega} u v + \int_{\Omega} \nabla u \nabla v$ and induced norm $||u||_{H^1} = ||u||_{L^2} + ||\nabla u||_{L^2} $.  \\[5pt]
$H^1_0(\Omega)$ & \quad & completion of $C^{\infty}_c(\Omega)$ in the norm $||u||_{H^1}$. If $\Omega$ is bounded, we have the equivalent norm given by $||u||_{H^1} = ||\nabla u||_{L^2} $.\\[5pt]
$C^{\infty}_c(\Omega)$ & \quad & space of smooth functions $u: \Omega \to \mathbb{R}$ that have compact support in $\Omega$. \\

\bottomrule
\end{tabular}
\end{center}
\label{tab:TableOfNotations}
\end{table}

\subsection{The Darcy flow equation.}\label{darcy_example} Here we present a case where using our method of generating data does not work very well compared to using the dataset provided in \cite{NeuralOperator}. The Darcy flow equation is given by
\begin{equation}\label{darcy}
    \left\{\begin{aligned}
-\operatorname{div}(a(x) \cdot \nabla u)&=f && \text{in } \Omega\\[5pt]
B(u) & = 0 && \text{on } \partial{\Omega},
\end{aligned}\right.
\end{equation}

In the paper \cite{NeuralOperator}, they fix $f \equiv 1$ and they are interested in learning the operator mapping the coefficients $\alpha$ into the solution $u$
\begin{align*}
    G^\dagger: L^{\infty}(\Omega) &\longrightarrow H_0^1(\Omega) \\
    \alpha & \longmapsto u
\end{align*}
In our setting, we are trying to learn the operator mapping the coefficients in $\alpha$ and the forcing term $f$ into the solution $u$
\begin{align*}
    G: L^{\infty}(\Omega) \times L^2(\Omega) &\longrightarrow H_0^1(\Omega) \\
    (\alpha,f) & \longmapsto u
\end{align*}
Here, $\alpha \sim \mu$ where $\mu $ is the pushforward of a Gaussian measure with covariance $C=(-\Delta +9I)^{-2}$ under the map
\begin{align*}
    T: \mathbb{R} &\longrightarrow \mathbb{R}_+ \\
    x & \longmapsto \begin{cases}
    12, & x \geq 0 \\
    3, & x \le 0
    \end{cases}
\end{align*}
Notice that by construction the coefficients are not smooth. We make some slight modifications to the FNO architecture so that it can take two functions $(\alpha,f)$ as an input and train FNO with $100,000$ data points. For testing we use functions from the FNO dataset on the Darcy flow and and passing the input in the form $(\alpha, 1)$. Below we summarize performance of FNO trained with our data while testing is done with the FNO dataset. However, for examples of problems where the coefficients $\alpha$ are smoother, our method generalizes better.

\begin{figure}[!ht]

\begin{center}
\centerline{\includegraphics[width=\columnwidth]{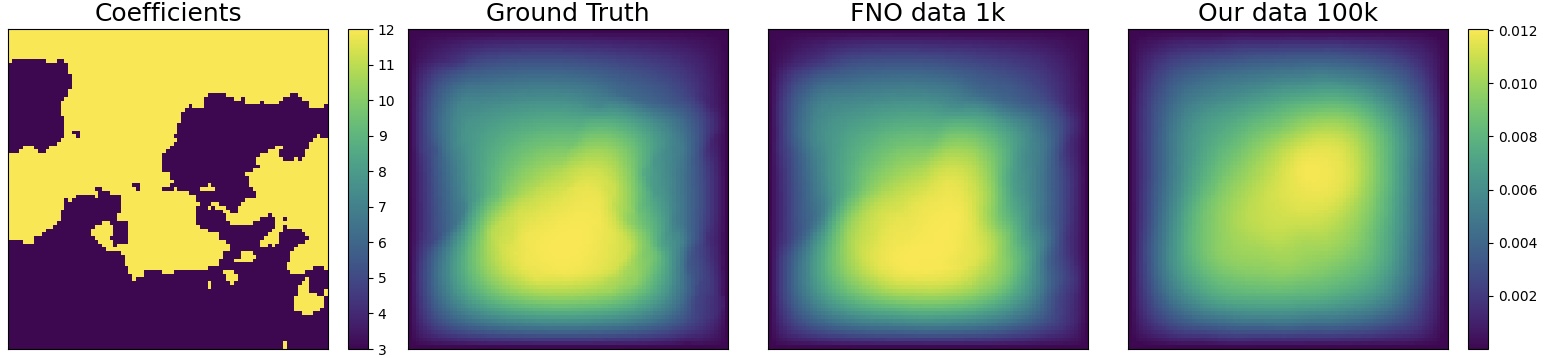}}
\caption{Predicted solutions of the Darcy flow equation using FNO with $1,000$ of the FNO data set and training with $100,000$ of our training data. The relative $L^2$ errors are respectively $0.012$ and  $0.174$.}
\label{fig3a}
\end{center}

\begin{center}
\centering{\includegraphics[width=\columnwidth]{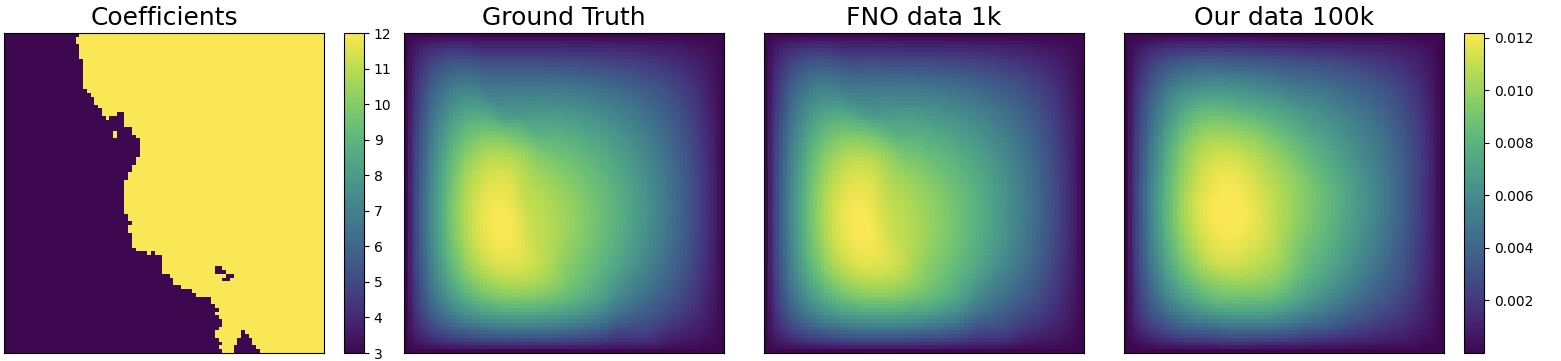}}
\caption{Predicted solutions of the Darcy flow equation using FNO with $1,000$ of the FNO data set and training with $100,000$ of our training data. The relative $L^2$ errors are respectively $0.005$ and  $0.071$.}
\label{fig3b}
\end{center}
\end{figure}

\end{document}